\documentclass[runningheads]{llncs}

\usepackage{graphicx}
\usepackage{amsmath,amssymb,amsfonts,nicefrac,derivative}
\usepackage[title]{appendix}
\usepackage{csquotes}
\usepackage{caption}
\usepackage{subcaption}
\usepackage[inline]{enumitem}
\usepackage{booktabs}
\usepackage{algorithm}
\usepackage{tabularx}
\usepackage{algpseudocode}
\algdef{SE}[DOWHILE]{Do}{doWhile}{\algorithmicdo}[1]{\algorithmicwhile\ #1}%

\usepackage{listings}
\usepackage{tikz}
\tikzset{>=latex}

\usepackage{epstopdf}

\usepackage{url}
\usepackage{cleveref}
\crefname{section}{Sect.}{Sect.}
\Crefname{section}{Section}{Sections}
\crefname{figure}{Fig.}{Fig.}
\crefname{algorithm}{Algorithm}{Algorithm}
\crefname{appendix}{Appendix}{Appendix}
\crefname{table}{Table}{Table}

\usepackage[acronym]{glossaries}
\usepackage{glossaries-extra}

\newcommand{\nR}{\mathbb{R}}
\newcommand{\dif}[1]{\mathrm{d}{#1}\,}
\newcommand{\CK}{\text{CK}}
\newcommand{\ckmin}{\text{ckmin}}
\newcommand{\strain}{\text{strain}}

\DeclareMathOperator*{\argmin}{arg\,min}

\setabbreviationstyle[acronym]{long-short}
\newacronym{ai}{AI}{Artificial Intelligence}
\newacronym{api}{API}{Application Programming Interface}
\newacronym{ml}{ML}{Machine Learning}
\newacronym{ann}{ANN}{Artificial Neural Network}
\newacronym{pp}{PP}{Piecewise Polynomial}
\newacronym{pinn}{PINN}{Physics-informed Neural Network}
\newacronym{rl}{RL}{Reinforcement Learning}

\begin{document}
\title{ Machine Learning Optimized Orthogonal Basis Piecewise Polynomial
  Approximation%
  \thanks{This work was supported by the Christian Doppler Research
  Association (JRC ISIA), the federal state of Salzburg WISS-FH project IAI and
  the European Interreg Österreich-Bayern project BA0100172 AI4GREEN. This
  preprint has not undergone peer review or any post-submission improvements or
  corrections.} 
  }
\titlerunning{ML Optimized Orthogonal Basis PP Approximation}

%
%
\author{Hannes Waclawek \inst{1}, \inst{2} \and Stefan Huber \inst{1}}
\authorrunning{H. Waclawek \and S. Huber}
%
%

\author{Hannes Waclawek \inst{1,2} \and Stefan Huber \inst{1}}

\institute{Josef Ressel Centre for Intelligent and Secure Industrial
Automation,\\Salzburg University of Applied Sciences, Austria \and Paris Lodron
University Salzburg, Austria\\
\email{\{hannes.waclawek, stefan.huber\}@fh-salzburg.ac.at}}

\maketitle              

\begin{abstract}
  \glspl{pp} are utilized in several engineering disciplines, like trajectory
  planning, to approximate position profiles given in the form of a set of
  points. While the approximation target along with domain-specific
  requirements, like $\mathcal{C}^k$-continuity, can be formulated as a system
  of equations and a result can be computed directly, such closed-form solutions
  posses limited flexibility with respect to polynomial degrees, polynomial
  bases or adding further domain-specific requirements. Sufficiently complex
  optimization goals soon call for the use of numerical methods, like gradient
  descent. Since gradient descent lies at the heart of training \glspl{ann},
  modern \gls{ml} frameworks like TensorFlow come with a set of gradient-based
  optimizers potentially suitable for a wide range of optimization problems
  beyond the training task for \glspl{ann}. Our approach is to utilize the
  versatility of \gls{pp} models and combine it with the potential of modern
  \gls{ml} optimizers for the use in function approximation in 1D trajectory
  planning in the context of electronic cam design. We utilize available
  optimizers of the \gls{ml} framework TensorFlow directly, outside of the scope
  of \glspl{ann}, to optimize model parameters of our \gls{pp} model. In this
  paper, we show how an orthogonal polynomial basis contributes to improving
  approximation and continuity optimization performance. Utilizing Chebyshev
  polynomials of the first kind, we develop a novel regularization approach
  enabling clearly improved convergence behavior. We show that, using this
  regularization approach, Chebyshev basis performs better than power basis for
  all relevant optimizers in the combined approximation and continuity
  optimization setting and demonstrate usability of the presented approach
  within the electronic cam domain.
  \keywords{Piecewise Polynomials \and Gradient Descent \and Chebyshev Polynomials 
  \and Approximation \and TensorFlow \and Electronic Cams}
\end{abstract}

\section{Introduction}
\subsection{Motivation} \label{sec:motivation}
\glspl{pp} are of special interest in several science and engineering
disciplines, where the latter are particularly interesting, as they come
with additional physical constraints. Path and trajectory planning for machines
in the field of mechatronics are just two examples. Path planning is the task of
finding possible waypoints of a robot or automated machine to move through its
environment without collision. Trajectory planning computes time-dependent
positional, velocity or acceleration profiles that hold setpoints for
controllers of robots or automated machines to move joints from one waypoint to
the other. Electronic cams are a subfield of the latter, describing repetitive
motion executed by servo drives within industrial machines. In this context,
positional, velocity or acceleration profiles are defined as input point clouds
and approximated by \glspl{pp}, which are then processed by industrial servo
drives, like B\&R Industrial Automation's ACOPOS series. Conventionally, the
approximation target along with domain-specific requirements such as continuity,
cyclicity or periodicity are formulated as a system of equations and a
closed-form solution is computed. While computational effort may be low for
computing such closed-form solutions, they posses limited flexibility with
respect to polynomial degrees, polynomial bases or adding further
domain-specific requirements. Sufficiently complex optimization goals soon call
for the use of numerical methods, like gradient descent.

Training an \gls{ann} is the iterative process of adapting weights of node
connections in order to fit given training data. With the advent of \gls{ml}
frameworks, this iterative training task is commonly carried out by calculating
gradients using automatic differentiation and updating weights according to the
gradient descent algorithm. Since gradient descent lies at the heart of this
process, modern \gls{ml} frameworks like TensorFlow or PyTorch therefore come
with a set of gradient-based optimizers potentially suitable for a wide range of
optimization problems beyond the training task for \glspl{ann}. Since advances
in deep learning rely on these optimizers, state-of-the art insights within the
field are continuously implemented, holding the potential of better performance
than classical gradient descent optimizers.

Our approach therefore is to utilize the versatility of \gls{pp} models and
combine it with the potential of modern \gls{ml} optimizers for the use in
function approximation in 1D trajectory planning in the context of electronic
cam design. We utilize available optimizers of the \gls{ml} framework TensorFlow
directly, outside of the scope of \glspl{ann}, to optimize model parameters of
our \gls{pp} model. This allows for an optimization of trajectories with respect
to dynamic criteria using state of the art methods, while at the same time
working with an explainable model, allowing to directly derive physical
properties like velocity, acceleration or jerk. Although our work documented in
\cite{huber2023} shows that our approach is basically feasible, experiments show
that clean convergence especially regarding continuity optimization requires
further practical considerations and regularization techniques. In an attempt to
overcome these shortcomings, in this paper, we show how an orthogonal polynomial
basis contributes to significantly improving approximation and continuity
optimization performance.


\subsection{An Orthogonal Basis}

Orthogonal polynomial sets are beneficial for determining the best polynomial
approximation to an arbitrary function $f$ in the least squares sense, resulting
from the fact that every polynomial in such an orthogonal system possesses a
minimal $L_2$ property with respect to the system's weight function, see
\cite{mason2002} for details. Among different sets of orthogonal polynomials,
Chebyshev polynomials of the first kind are of special interest. Besides
favorable numerical properties, like the absolute values such polynomials
evaluate to staying within the unit interval, their roots are well known as
nodes (Chebyshev nodes) in polynomial interpolation to minimize the problem of
Runge's phenomenon. This effect is interesting for us, as we are anticipating
reduced oscillating behavior beneficial for our electronic cam approximation
setting. Additionally and contrary to Chebyshev polynomials of the second kind,
numerous efficient algorithms exist for the conversion between orthogonal basis
and power basis representations \cite{bostan2010}. (We need this in the end,
since our solution targets servo drives like B\&R Industrial Automation's ACOPOS
series processing cam profiles as a power basis \gls{pp} function.) With respect
to potential future work, the relationship of Chebyshev series to Fourier cosine
series has the potential of allowing a more straightforward formal approach to
spectral analysis and optimization of generated cam profiles (e.g.\ to reduce
resonance vibrations), since it inherits theorems and properties of Fourier
series, allowing for a Chebyshev transform analogue to Fourier transform
\cite{datta1995}. These arguments make Chebyshev polynomials of the first kind
an attractive subject of investigation for this paper.

\subsection{Related work} 
There is a lot of published work on approximation using \gls{pp} in the cam
approximation domain that rely on a closed-form expression in order to
non-iteratively calculate a solution, like \cite{mermelstein2004}. Although
computational effort is reduced using such approaches, as mentioned earlier,
they posses limited flexibility with respect to polynomial degrees, bases and
complexity of loss functions.
There also is work on \gls{pp} approximation using \enquote{classical}
gradient based approaches. In \cite{chen2014}, Voronoi tessellation is used for
partitioning the input space before approximating given 2D surfaces using a
gradient-based approach with multivariate \glspl{pp}. The authors state that an
orthogonal basis is beneficial for this task, however, develop their approach
using power basis. Contrary to our work, the loss function only comprises the
approximation error, which implies that continuity is not achieved by their approach.
Furthermore, the authors state that magnitudes of derivatives of the error
function may vary greatly, thus making classical gradient descent approaches
unusable. We show in this paper, that regularization may mitigate this effect,
at least for continuity optimization in the 1D setting.

Other work utilizes parametric functions for approximation of
given input data, like B-Spline or NURBS curves, as in \cite{nguyen2019}. This
has some benefits over non-parametric \glspl{pp}: For one,
$\mathcal{C}^k$-continuity is given simply by the choice of order of the curve.
For another, fewer parameters (usually control points) need to be optimized.
However, \gls{pp} models are utilized in several engineering disciplines and
conversion from parametric curves is not generally possible. As an example,
servo drives, like B\&R Industrial Automation's ACOPOS series, process
electronic cams in a \gls{pp} format.

\glspl{ann} are investigated for the use in function approximation. In
\cite{adcock2021}, the authors provide an overview of the state-of-the art with
respect to aspects like numerical stability or accuracy and develop a
computational framework for examining the performance of \glspl{ann} in this
respect. Although results look promising and other recent works, like
\cite{shen2020}, already provide algorithmic bounds for the \gls{ann}
approximation setting, one of the main drivers for the use of a \gls{pp} model
outside of the context of neural networks for our work is explainability. When
utilizing \gls{ai} methods in the domain of electronic cams, there is a strong
necessity for explainability, mainly rooted in the fact that movement of motors
and connected kinematic chains like robotic arms affects its physical
environment with the potential of causing harm, therefore calling for
predictable movement. This, in turn, calls for a \enquote{predictable model},
that allows to reliably derive physical properties and generate statements about
expected behavior. Another recent development tackling these challenges are
\glspl{pinn}, where resulting \glspl{ann} can be described using partial
differential equations. For the domain of 1D electronic cam approximation
covered in this work, however, using \gls{pp} models is simpler and more
intuitive considering that servo drives directly utilize \glspl{pp}.

There are some preliminary non-scientific texts (e.g.\ personal blog articles)
on gradient descent optimization for polynomial regression utilizing \gls{ml}
optimizers for a single polynomial segment. However, to the best of
our knowledge, there is no published work on utilizing gradient descent
optimizers of modern \gls{ml} frameworks for $\mathcal{C}^k$-continuous \gls{pp}
approximation other than our previous research published in \cite{huber2023}.

\subsection{Contributions}

Although Chebyshev basis holds the potential of improved approximation
optimization performance, our experiments show that clean convergence, especially
regarding continuity optimization, requires further practical considerations and
regularization techniques. In this work, we therefore
\begin{enumerate}
  \item adapt our base approach introduced in \cite{huber2023} to utilize an
  orthogonal basis using Chebyshev polynomials of the first kind,
  \item develop a novel regularization approach enabling clearly improved
  convergence behavior,
  \item show that, using this regularization approach, Chebyshev basis performs
  better than power basis for all relevant optimizers in the combined
  approximation and continuity optimization setting and
  \item demonstrate usability of the presented approach within the electronic cam
  domain by comparing remaining approximation and continuity losses
  to least squares optima and strictly establishing continuity via an algorithm
  with impact only on local polynomial segments.
\end{enumerate}

We do so by studying convergence behavior and properties of generated curves in
the context of controlled experiments utilizing the popular \gls{ml} framework
TensorFlow. All experimental results discussed in this work are available at 
\cite{waclawek2024}.

\section{Piecewise Polynomial Model} \label{sec:model}
Let $d \in \mathbb{N}_0$.
The Chebyshev polynomials of the first kind are defined via the recurrence
relation $2x\;T_{d-1}(x)-T_{d-2}(x)$, with $T_d(x) = 1$ if $d = 0$ and 
$T_d(x) = x$ for $d = 1$.
The Chebyshev polynomials form an orthogonal set of functions on the interval
$[-1,1]$ with respect to the weighting function $w(x)=\frac{1}{\sqrt{1-x^2}}$.
This means that two Chebyshev polynomials $T_c(x)$ and $T_d(x)$ are orthogonal
with respect to the inner product $T_d(x) \rangle = \int_{-1}^1
T_c(x)T_d(x)\;\frac{dx}{\sqrt{1-x^2}}$.

Considering $n$ samples at $x_1 \le \dots \le x_n \in \nR$ with respective
values $y_i \in \nR$, we ask for a \gls{pp} $f \colon I \to \nR$ on the interval
$I = [x_1, x_n]$ approximating the input samples well and fulfilling additional
domain-specific properties, like $\mathcal{C}^k$-continuity. The polynomial
boundaries of $f$ are denoted by $\xi_0 \le \dots \le \xi_m$ , where $\xi_0 =
x_1$ and $\xi_m = x_n$. With $I_i = [\xi_{i-1}, \xi_i]$, the \gls{pp} $f$ is
modeled by $m$ polynomials $p_i \colon I \to \nR$ that agree with $f$ on $I_i$
for $1 \le i \le m$. The Chebyshev polynomials up to $d$ form a basis of the
vector space of real-valued polynomials up to degree $d$ in the interval $[-1,
1]$. This means that an arbitrary real-valued polynomial $p(x)$ up to degree $d$
defined on the interval $[-1, 1]$ can be expressed as a linear combination of
Chebyshev polynomials of the first kind. We therefore can adapt our \gls{pp}
model introduced in \cite{huber2023} and construct each polynomial segment $p_i$
via Chebyshev polynomials of the first kind as
\begin{align}
  \label{eq:chebyshev_series_segment}
  p_i = \sum_{j=0}^d c_{i,j} T_j(x-\mu_i).
\end{align}
We rescale input data and shift polynomials to the mean of the respective
segment by $\mu_i = \frac{\xi_{i-1}+\xi_{i}}{2}$, as outlined in
\cref{sec:experiments}. The $c_{i,j}$ are the to be trained model parameters of
the \gls{pp} \gls{ml} model. We investigate the convergence of these model
parameters $c_{i,j}$ with respect to the loss function defined below.

\subsection{Loss function}  \label{sec:loss}
We utilize the loss function 
\begin{align}
  \label{eq:total_loss}
    \ell = \alpha \ell_{\CK} + (1-\alpha) \ell_2
\end{align}
with $0 \le \alpha \le 1$ in order to establish $\mathcal{C}^k$-continuity and
allow for curve fitting via least squares approximation, where the approximation
error $\ell_2$ is defined as
\begin{align}
  \label{eq:l2_loss}
  \ell_2 = \frac{1}{n} \sum_i |f(x_i) - y_i|^2.
\end{align}
Note that, as outlined in \cref{sec:experiments_ell}, optimization results with
$\alpha = 0$ are of less practical relevance, because remaining approximation
errors are significantly higher after strictly establishing continuity utilizing
the algorithm introduced in \cref{sec:continuity_algorithm}. Summing up
discontinuities at all $\xi_i$ across relevant derivatives as
\begin{align}
  \label{eq:ck_loss}
  \ell_\CK = \frac{1}{m - 1} \sum_{i=1}^{m-1} \sum_{j=0}^k \left(\frac{\Delta_{i,j}}{r_k}\right)^2
  \quad \text{with} \quad
  \Delta_{i,j} = p^{(j)}_{i+1}(\xi_i) - p^{(j)}_i(\xi_i)
\end{align}
quantifies the amount of discontinuity throughout the overall \gls{pp} function.
This loss definition allows for a straightforward extension towards
$\mathcal{C}^k$-cyclicity or periodicity commonly required for electronic cam
profiles: We can achieve periodicity in \eqref{eq:ck_loss} by adapting $m - 1$
to $m$ and generalizing $\delta_{i,j} = p^{(j)}_{1 + (i \bmod m)}(\xi_{i \bmod
m}) - p^{(j)}_i(\xi_i)$. For cyclicity, we ignore the case $j=0$ when
$i=m$. (While periodicity requires matching the values of all derivatives at the
endpoints of a motion profile, with cyclicity, we can have a positional offset.)
For the sake of generality, we define $\ell_\CK = 0$ for $m=1$.

Experiments documented in \cref{sec:experiments} show that the derivative-specific
regularization factor $r_k$ is required in order to reduce
oscillating behavior and improve convergence behavior. In the following, we 
develop a formula describing this derivative-specific
regularization factor.

\subsection{Regularization of Loss} \label{sec:regularization}

Magnitudes of discontinuities at boundary points may vary significantly between
derivatives, with higher derivatives potentially having higher magnitudes, thus
impairing convergence. With polynomials, the potentially highest contributing
factor is the one of the highest order term. We can express a derivative $k$ of
a polynomial of degree $d$ in power basis form using Horner's method as $p_i =
\sum_{i=k}^d c_{i} x^{i-k} \cdot R_k$, where $R_k = (i \cdot (i-1) \cdot (i-2)
\cdot ... \cdot (i-k-1))$.
For the highest order term, this leaves us with
\begin{align}
  \label{eq:regularization_factor}
  r_k &= \frac{d!}{(d-k)!}.
\end{align}
Our approach is to regularize each derivative by $r_k$ defined in
\cref{eq:regularization_factor} as outlined in \cref{eq:ck_loss}. We could apply
a similar approach to Chebyshev basis. Since, for calculating $\ell_{CK}$, we
are only interested in functional values at the boundaries of individual
polynomial segments, due to symmetry properties of Chebyshev polynomials, we
receive the same absolute values at both ends of the interval $[-1, 1]$. We
therefore don't need a closed formula also for Chebyshev basis, but could simply
evaluate $\|T_d(1)\|$ in order to retrieve the regularization factor for
derivative $d$. However, experiments indicate that $r_k$ defined in
\cref{eq:regularization_factor} performs better also for Chebyshev basis. We
therefore utilize this factor for regularization of continuity loss $\ell_{\CK}$
with both polynomial bases.

\subsection{Enforcing Continuity} \label{sec:continuity_algorithm}

We can eliminate possible remaining discontinuities, i.e., $\ell_{\CK} > 0$ in
the sense of \cref{eq:ck_loss}, by applying corrective polynomials that strictly
enforce $\Delta(\xi_i) = 0$ at all $\xi_i$. Iterating through polynomial
segments from left to right, we construct a corrective polynomial $q_i$ of
degree $2k+1$ by formulating a system of equations taking the values of $k+1$
derivatives per endpoint, respectively, as shown in \cref{alg:continuity} in
\cref{sec:appendix_alg}. Note that the corrective polynomials themselves are of
degree at most $d$.

A favorable property of \cref{alg:continuity} is that it only has local effects:
only derivatives relevant for $\mathcal{C}^k$-continuity are modified, while
other derivatives remain unchanged. This allows us to apply the corrections at
each $\xi_i$ independently as they have only local impact. This is a nice
property in contrast to interpolation methods like natural cubic splines or
parametric curves lacking the local control property, like Bézier curves. For an
extension towards $\mathcal{C}^k$-cyclicity or periodicity commonly required for
electronic cam profiles, we can naturally modify the first cases of $b^L$ and
$b^R$ of \cref{alg:continuity} as outlined in \cref{sec:loss}, respectively.

\subsection{Discussion of Continuity Loss Characteristics} \label{sec:non-convexity}

Let us denote by $\theta$ the list of model parameters $c_{1,0},
c_{1,1}, \dots, c_{m,d}$ and let us denote by $f_\theta$ the
resulting \gls{pp}.
Using the mean-square-error loss function outlined in \cref{eq:l2_loss},
a natural approach would be to solve the optimization problem
\begin{align}
  \label{eq:ell2problem}
  \theta^* = \argmin_{\theta} \ell_2,
\end{align}
to obtain the optimal model parameters $\theta^*$. Since this problem is convex,
there is a unique solution. However, $f_{\theta^*}$ is in general not
$\mathcal{C}^k$-continuous, or even $\mathcal{C}^0$-continuous, as the
individual $p_i$ are optimized individually.

We can capture the amount by which $\mathcal{C}^k$-continuity is violated by
the $\ell_\CK$ loss defined in \cref{eq:ck_loss}.
 Note that
$\Delta_{i,j}$ captures the discontinuity of the $j$-th derivative at $\xi_i$.
Also note that $\ell_\CK = 0$ iff $\mathcal{C}^k$-continuity holds.

Then we ask for the $\ell_2$-minimizing $\mathcal{C}^k$-continuous
\gls{pp}, which is the solution of the constrained optimization problem
\begin{align}
  \label{eq:constrainedproblem}
  \begin{aligned}
    \min_{\theta} \quad& \ell_2\\
    \textrm{s.t.} \quad& \ell_\CK = 0.
  \end{aligned}
\end{align}
We turn this into an unconstrained optimization problem by adding $\ell_\CK$ to
the loss function in \cref{eq:total_loss} and obtain
\begin{align}
  \label{eq:theproblem}
  \begin{aligned}
    \theta^*_\alpha = \argmin_{\theta} \ell
  \end{aligned}
\end{align}
with $0 \le \alpha \le 1$.
For $\alpha = 0$, we again have the problem in \cref{eq:ell2problem}. If $\alpha
= 1$ then any $\mathcal{C}^k$-continuous \gls{pp} is an optimal solution, such
as the zero function, but also the solution of \cref{eq:constrainedproblem}. Any
$\alpha > 0$ leaves us, in general, with a non-convex optimization problem prone
of getting stuck in local optima. Let us denote by $f^*_\alpha$ a solution of
\cref{eq:theproblem} with a fixed $\alpha$. In this sense, we are interested in
a $f^*_\alpha$ approximating the input point set well after running the
algorithm introduced in \cref{sec:continuity_algorithm}, since the result is
guaranteed to be numerically continuous. In \cref{sec:experiments} we therefore
perform experiments analyzing suitable initialization methods and $\alpha$
values for different input point sets and preconditions, with a subsequent
strict establishment of continuity via \cref{alg:continuity}.

\section{Experimental Results} \label{sec:experiments}

The Chebyshev polynomials up to $d$ form a basis of the vector space of
real-valued polynomials up to degree $d$ in the interval $[-1, 1]$. We therefore
rescale the $x$-axis of input data so that every segment is of span $2$.
Considering this, for a \gls{pp} consisting of $4$ polynomial segments, as an
example, we scale the input data such that $I = [0, 8]$. Splitting the interval
equally into $4$ segments we receive a width of $2$ for each individual segment,
and, by $\mu_i$ in \cref{eq:chebyshev_series_segment}, we shift polynomials to
the mean of the respective segment, leading to $I_i = [-1, 1]$ for each
respective \gls{pp} segment. We do the same for power basis and skip
back-transformation as we would do in production code. (In our experiments,
convergence is not significantly impaired for interval widths of $2 \pm 0.5$ for
Chebyshev basis with tested data sets.)

Considering our focus on the electronic cam domain, a position profile with high
acceleration values leads to the kinematic system being exposed to high forces.
A position profile with high jerk values, on the other hand, will make the
system prone to vibration and excessive wear \cite{josephs2002}. In order to
address the latter, continuos jerk curves are highly beneficial, if not a
prerequisite. We therefore look closer at $\mathcal{C}^3$-continuity in all
conducted experiments with $\alpha > 0$. Looking at \cref{alg:continuity}, this
implies a \gls{pp} of degree $7$. We utilize the Tensorflow
\textit{GradientTape} environment for the creation of custom training loops
utilizing available optimizers directly, as outlined in \cite{huber2023}. All
experimental results discussed in this section where created with TensorFlow
version $2.13.0$, Python $3.10$ and are available at \cite{waclawek2024}.

\subsection{Input Data and Optimization Goal} \label{sec:experiments_input_goal}

Computing \cref{eq:ell2problem} for each polynomial segment, we denote by 
\begin{align}
  \label{eq:ell_2_star}
  \begin{aligned}
    \ell_2^* = \min_{\theta} \quad& \ell_2
  \end{aligned}
\end{align}
the segment-wise approximation loss optimum. The resulting \gls{pp} is not
$C^k$-continuous, since each polynomial segment is fitted individually.
Utilizing method CKMIN described in \cref{alg:continuity}, we enforce continuity
for this result and denote by $\widetilde{\ell_2^*} =
\ell_2(\ckmin(\argmin_{\theta}\ell_2))$ its approximation loss. It is easy to
see that $\ell_2^* \leq \widetilde{\ell_2^*}$.  Considering this, our
optimization goal is for results to lie in the margin between $\ell_2^*$ and
$\widetilde{\ell_2^*}$. The closer a result is to $\ell_2^*$, the better. The
value of $\ell_2^*$ depends on the amount of variance / noise in the input data
along with the choice of number of polynomial segments. In the following, we
therefore refer to the value of $\ell_2^*$ as \textit{variance}, i.e., input
data with a higher value of $\ell_2^*$ is referred to as input data with higher
variance. \cref{tab:input} gives an overview of input data used for our
experiments along with the respective baseline values of $\ell_2^*$ and
$\widetilde{\ell_2^*}$ when enforcing $\mathcal{C}^3$-continuity. In addition to
the input data described in \cref{tab:input}, we utilize two different noise
levels for each dataset, generated by adding random samples from a normal
Gaussian distribution using the package \texttt{numpy.normal} with scales $0.1$
and $0.5$, respectively, and a fixed seed of $0$ for reproducibility.

\setlength{\tabcolsep}{12pt}
\begin{table}[htb]
  \centering
  \begin{tabularx}{\linewidth}{@{}Xlllll@{}}
  \toprule
  Dataset & $I$ & $n$ & $m$ &
  $\ell_2^*$ & $\widetilde{\ell_2^*}$ \\ \midrule A: $\sin(x)$ & $[0, \frac{\pi}{2}]$
  & $50$ & $2$ & $4.110 \cdot 10^{-21}$ & $8.450 \cdot 10^{-17}$ \\
  B: $\sin(x)$ & $[0, 2 \pi]$ & $100$ & $2$ & $2.240 \cdot 10^{-11}$ & $3.630
  \cdot 10^{-7}$ \\
  C: $\sin(x^2 4 \pi)$ & $[0, 1]$ & $100$ & $3$ & $3.540 \cdot 10^{-6}$ & $4.230
  \cdot 10^{-2}$ \\ \bottomrule
  \end{tabularx}
  \caption{Different input data sets used for performing experiments.}
  \label{tab:input}
\end{table}

\subsection{Optimizing $\ell_2$ only} \label{sec:experiments_ell2}

\begin{figure}[htb]
  \centering
  \includegraphics[width=\textwidth]{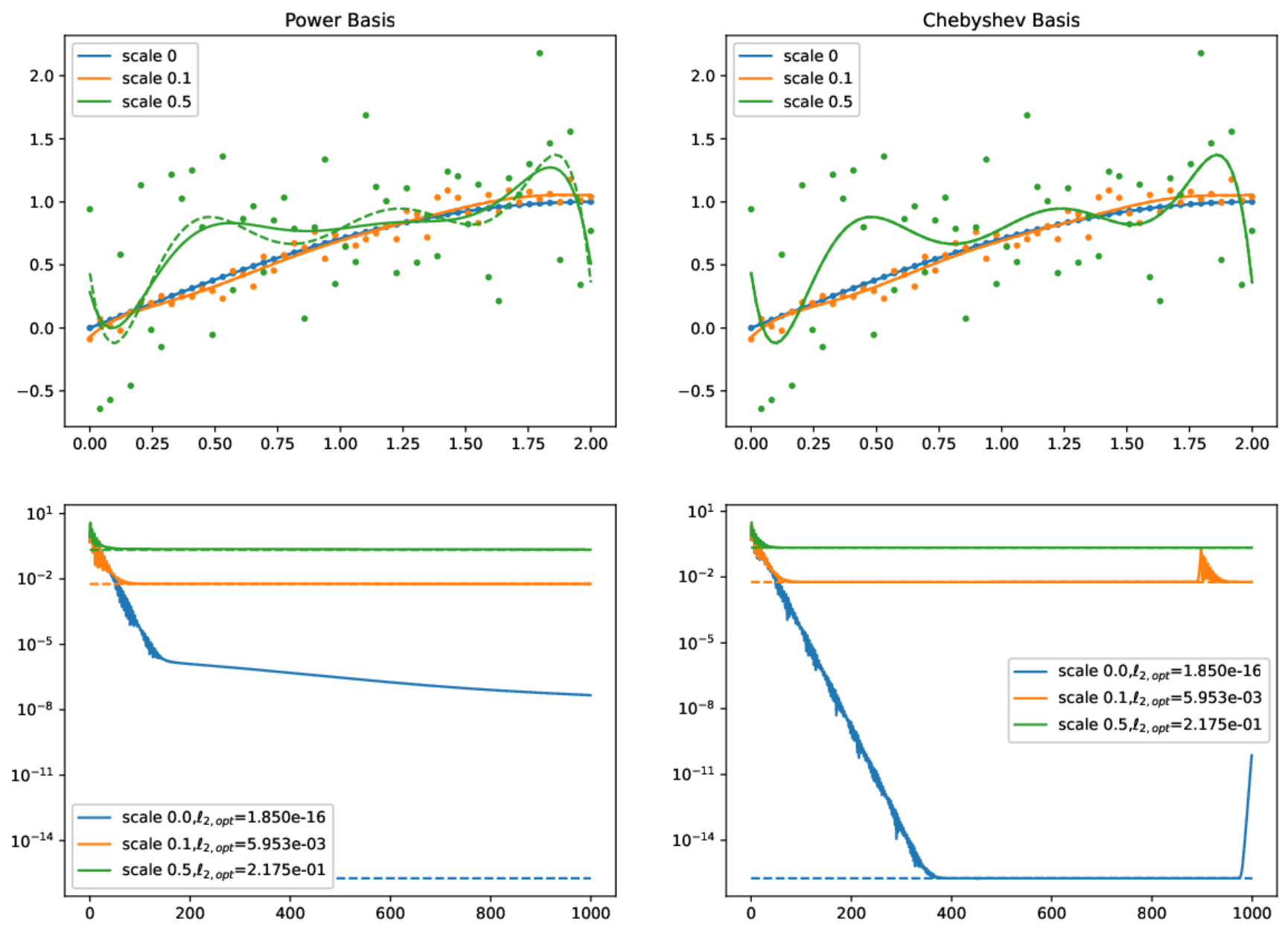}
  \caption{Results for different noise levels, AMSGrad optimizer, learning rate
  $1.0$, dataset A, $1$ segment, $1000$ epochs, $\alpha = 0$. Top: Derivative
  $0$ curve shapes. Bottom: $\ell_2$ losses. Optima at dashed lines.}
  \label{fig:noise_l2_only}
\end{figure}

We first compare performance of Chebyshev basis and power basis in the single
approximation target scenario, i.e., one polynomial segment and $\alpha = 0$.
With rising polynomial degree, the remaining $\ell_2$ optimum approximation
error gets lower. This is in accordance with Chebyshev basis performance
observed in our experiments, where results manage to converge to a lower loss
with rising degree.  
This is not the case for power basis. While performance with degree $5$ is
competitive, higher degrees perform clearly worse than the Chebyshev basis
counterparts. As expected, higher polynomial degrees generally require more
epochs to converge for both bases.  
For Chebyshev basis, however, results generally converge to lower remaining
losses, and, for all observed degrees ($[3, 9]$), within $1000$ epochs.  
Experiments also show that a learning rate of $1.0$ is a reasonable choice for
all observed degrees and both observed polynomial bases in the sole
approximation optimization target scenario. Raising the number of input points
has no effect to the highest learning rate we can achieve.  
Of course, the number of input points has to be sufficiently high for a
polynomial degree to enable a well-conditioned fit in the first place. Looking
at the impact of the amount of variance in the input data, generally speaking,
it takes longer to converge to the optimal approximation result for both
observed polynomial bases with less noise in the input data. However, while
optimization with Chebyshev basis reaches the least squares optimum for all
observed noise levels, power basis is only competitive with noise added to the
input curve. The experimental test run documenting this behavior for dataset A
is depicted in \cref{fig:noise_l2_only}.

Looking at the performance of all optimizers available with TensorFlow version
$2.13.0$ shows that Chebyshev basis clearly outperforms power basis with all
observed datasets and optimizers converging to the optimum within $5000$ epochs.
While none of the optimizers manage to reach the $\ell_2$ optimum with power
basis in the given $5000$ epochs, there are several optimizers achieving this
with Chebyshev basis. Interestingly, with Chebyshev basis, Vanilla SGD is
outperforming more \enquote{elaborate} adaptive optimizers, like AMSGrad and
other Adam-based optimizers. Sorted by quickest convergence, candidate
optimizers are Nadam, Adagrad, FTRL, Vanilla SGD, SGD with Nesterov momentum,
Adam, Adamax, AMSGrad SGD with momentum and Adadelta. As our main goal is
performing combined approximation and continuity optimization, we do not look
into optimizing hyperparameters of individual optimizers in the sole
approximation optimization target scenario. 

\subsection{Optimizing $\ell$} \label{sec:experiments_ell}

Contrary to the single approximation target discussed in the previous section,
we see that a lower learning rate of $0.1$ is beneficial. Chebyshev basis
results again converge to lower losses, however, now $2000$ epochs are required.
Increasing the number of segments does not affect the highest possible learning
rates.

Looking at the performance of all optimizers available with TensorFlow version
$2.13.0$, Chebyshev basis is again clearly outperforming power basis in regard
to all relevant optimizers and input point sets. \cref{fig:optimizers} in
\cref{sec:appendix} shows this in more detail for dataset A: While none of the
optimizers manage to surpass $\ell_2^*$ with power basis in the given $5000$
epochs, there are several optimizers achieving this with Chebyshev basis.
Interestingly, however, with Chebyshev basis, SGD does not converge in this
combined approximation and continuity optimization target any longer.
Results with regularization of $\mathcal{C}^k$-loss, as
introduced in \cref{sec:regularization}, clearly outperform results without
$\mathcal{C}^k$-loss regularization for candidate optimizers.

Running experiments with all datasets and noise levels defined in
\cref{tab:input}, we see the trend that more optimizers (also with power basis)
manage to surpass the $\ell_2^*$ baseline for input data with higher $\ell_2^*$
values. Experiments also clearly show that results with regularization of
$\mathcal{C}^k$-loss outperform results without $\mathcal{C}^k$-loss
regularization for candidate optimizers also for datasets other than dataset A.
The higher $\ell_2^*$ of the input data, the closer is the gap between
regularized and non-regularized losses, as well as power basis and Chebyshev
basis results. Considering all observed input data, AMSGrad with default
parameters is the best candidate for both polynomial bases in this combined
approximation and continuity optimization target. Since spikes in its total loss
curve frequently occur after the optimizer has reached lowest remaining losses,
however, early stopping with reverting to the best \gls{pp} coefficients
achieved for the training run is strongly recommended. Investigating different
methods of initializing polynomial coefficients ($\ell_2$ optimum, zero,
random), $\ell_2$ optimum initialization turns out to be the best choice. 

\begin{figure}[htb]
  \centering
  \includegraphics[width=\textwidth]{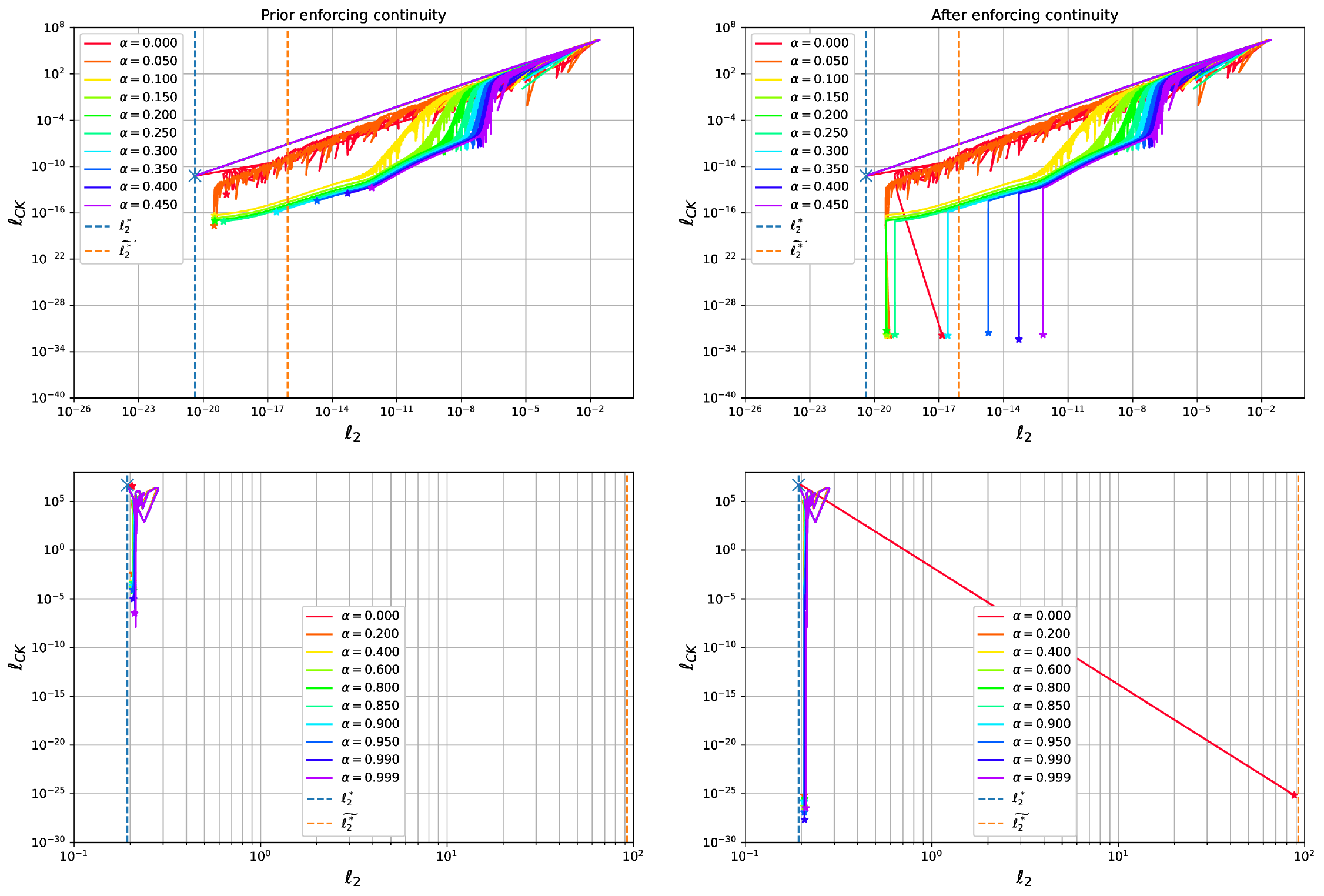}
  \caption{Continuity over approximation error for $\ell_2$ initialization
  with learning rate = $0.1$, $\alpha = 0.1$, dataset A, $2$ segments, $2000$
  epochs and early stopping enabled with a patience of $500$. Top: Noise scale = $0$. 
  Bottom: Noise scale = $0.5$.}
  \label{fig:lckoverl2}
\end{figure}

A simple approach for $\mathcal{C}^k$-continuous solutions would be to apply
\cref{alg:continuity} to the segment-wise $\ell_2$ optimum. However, looking at
$\ell_2$ versus $\ell_{CK}$ error for different datasets, as depicted in
\cref{fig:lckoverl2} for dataset A, we find that strictly establishing
continuity using \cref{alg:continuity} for a result with $\alpha = 0$ increases
the remaining approximation error significantly. Strictly establishing
continuity for an optimization result with $\alpha >0$, on the other hand,
affects prior approximation errors only mildly. This is an important finding, as
it not only substantiates the need for $\ell_{CK}$ optimization, but also tells
us that if we strictly establish continuity after optimization, it is beneficial
to choose an $\alpha$ value that is greater but close to $0$. 

Looking at the experimental setup of \cref{fig:lckoverl2}, the margin between
remaining approximation error $\ell_2^*$ of a conventional segment-wise least
squares fit and approximation error $\widetilde{\ell_2^*}$ of this fit after
strictly establishing continuity via algorithm is higher for input data with
higher variance. Taking this margin as a baseline, higher variance input data
leaves us with with a wider band of where our optimization leads to better
results than performing a conventional segment-wise least squares fit with
strict continuity establishment. In case of a wider band, we can also have a
wider range of $\alpha$ values that make sense. Leaving out some scenarios where
local optima of higher $\alpha$ values have a lower approximation error than
lower ones, we generally see a tendency of rising approximation errors with
rising $\alpha$ values. Considering the experimental setup of
\cref{fig:lckoverl2}, the two bottom plots show results for input dataset A with
a noise scale of $0.5$. Here, we observe results that lie within the
aforementioned margin throughout the complete $\alpha$ range $0 \leq \alpha \leq
1$. Without noise, as shown in the two top plots, the $\ell_2^*$ value of the
input data is lower, and we see $\alpha$ values of $> \approx 0.3$ tendentially
having ever higher approximation errors, thus leaving the margin, which leaves
us with a smaller range of possible values of $\alpha$ that make sense.

\section{Conclusion and Outlook}

Our results show that optimization of \glspl{pp} utlizing Chebyshev basis
clearly outperforms power basis with tested data sets and various noise levels.
Increasing the number of polynomial segments can further boost this advantage,
as Chebyshev basis benefits from input data with lower variance. Considering its
better performance with degree $7$ polynomials, this also makes Chebyshev basis
the favorable candidate for the generation of $\mathcal{C}^3$-continuos \gls{pp}
position profiles for the use in electronic cam approximation, as a polynomial
degree of $7$ is required to strictly establish $\mathcal{C}^3$-continuity after
optimization using \cref{alg:continuity}. Looking at Chebyshev basis convergence
results for different data sets, we see that the regularization method
introduced in \cref{sec:regularization} is required to reduce oscillating
behavior and boost optimizer performance. 

As outlined in section \cref{sec:non-convexity}, setting $\alpha > 0$ leaves us with
a non-convex optimization problem prone of getting stuck in unfavorable local
optima. Experimental results documented in \cref{sec:experiments} show, however,
that we can effectively guard against this by
\begin{enumerate*}[label=(\roman*)]
  \item pre-initializing polynomial coefficients with the segment-wise $\ell_2$
  optimum,
  \item applying early stopping and reverting to best \gls{pp} coefficients
  achieved during the training run and
  \item strictly establishing continuity after optimization with $\alpha > 0$
  running the algorithm introduced in \cref{sec:continuity_algorithm}.
\end{enumerate*}
Our results show that strictly establishing continuity for a result with $\alpha
= 0$ increases the remaining approximation error significantly, while a result
with $\alpha >0$ is only affected mildly. This substantiates the need for
$\ell_{CK}$ optimization also when strictly establishing continuity after
optimization. 

Looking at possible future work, additional terms in $\ell$ can accommodate for
further domain-specific goals. For instance, we can reduce oscillations in $f$
by penalizing the strain energy $\ell_{\strain} = \int_I f''(x)^2 \; \dif x$,
which, in the context of electronic cams, can reduce induced forces and energy
consumption. While our approach benefits from optimizers provided with modern
\gls{ml} frameworks, it also has the potential of contributing back to the field
of \glspl{ann}. In their preprint \cite{daws2019}, Daws and Webster initialize
Neural Networks via polynomial approximation using Legendre basis and show that
subsequent training results benefit from such an initialization. Our approach
could be used for such an initialization. Also, our approach is compatible to
other iterative \gls{ml} methods. In this way, we can utilize it for \gls{rl}
methods based on policy gradients, like Trust Region Policy Optimization or
Proximal Policy Optimization, and model the policy landscape using multivariate
polynomials in an effort of improving sample efficiency of modern \gls{rl}
algorithms. 

\newpage

\bibliographystyle{splncs04}
\bibliography{references}

\newpage

\appendix

\begin{subappendices}
  \renewcommand{\thesection}{\Alph{section}}%

  \section{Strictly Establishing Continuity} \label{sec:appendix_alg}

  \begin{algorithm}[h!]
    \caption{CKMIN: Strictly Establish $\mathcal{C}^k$-Continuity}
    \label{alg:continuity}
    \textbf{Input:} $PP, k$ \Comment{Piecewise polynomial, continuity class}
    \begin{algorithmic}[1]
    \Procedure{CKMIN()}{}
    \State $(p_1, \dots, p_m) \gets PP$ \Comment{$m$ polynomial pieces}
    \For{$i \gets 1$ \textbf{to} $m$} \Comment{Correct all $p_i$ over $[\xi_{i-1}, \xi_{i}]$}

      \State $A^L \gets \left( \frac{j!}{(l - j)!} \xi_{i-1}^{l-j} \right)_{j,l=0,j}^{k, 2k+1} $ \Comment{Left conditions at $\xi_{i-1}$}
      \State $A^R \gets \left( \frac{j!}{(l - j)!} \xi_{i}^{l-j} \right)_{j,l=0,j}^{k, 2k+1}$ \Comment{Right conditions at $\xi_{i}$}

      \State $b^L \gets \left(  
        \begin{cases}
          0 & \text{if } i = 0 \\ 
          p^{(j)}_{i-1}(\xi_{i-1}) -p^{(j)}_i(\xi_{i-1}) & \text{otherwise} 
        \end{cases} 
       \right)_{j=0}^{k} $ \Comment{Match $p_{i-1}$}
      
      \State $b^R \gets \left( 
        \begin{cases}
          0 & \text{if } i = m \\
          \frac{p^{(j)}_{i+1}(\xi_i) - p^{(j)}_i(\xi_i)}{2} & \text{otherwise}
           
        \end{cases} 
       \right)_{j=0}^{k} $ \Comment{Match mean of $p_{i}, p_{i+1}$}

      \State $A, b \gets \begin{pmatrix} A^L\\A^R \end{pmatrix}, \begin{pmatrix}b^L\\b^R \end{pmatrix}$ \Comment{Stacking equation systems}


      \State $c \gets A^{-1} \cdot b$ \Comment{Solve $A \cdot c = b$ for $c$}
      \State $q_i \gets \sum_{j=0}^{2k+1} c_{j} x^j$ \Comment{Corrective polynomial}
      \State $p_i \gets p_i + q_i$ \Comment{Apply correction}
    \EndFor
    \State \Return $(p_1, \dots, p_m)$ \Comment{$\mathcal{C}^k$-continuous piecewise polynomial}
    \EndProcedure
    \end{algorithmic}
  \end{algorithm}

  \newpage

  \section{Optimizer Performance} \label{sec:appendix}


  \begin{figure}[h!]
    \centering
    \includegraphics[width=\textwidth]{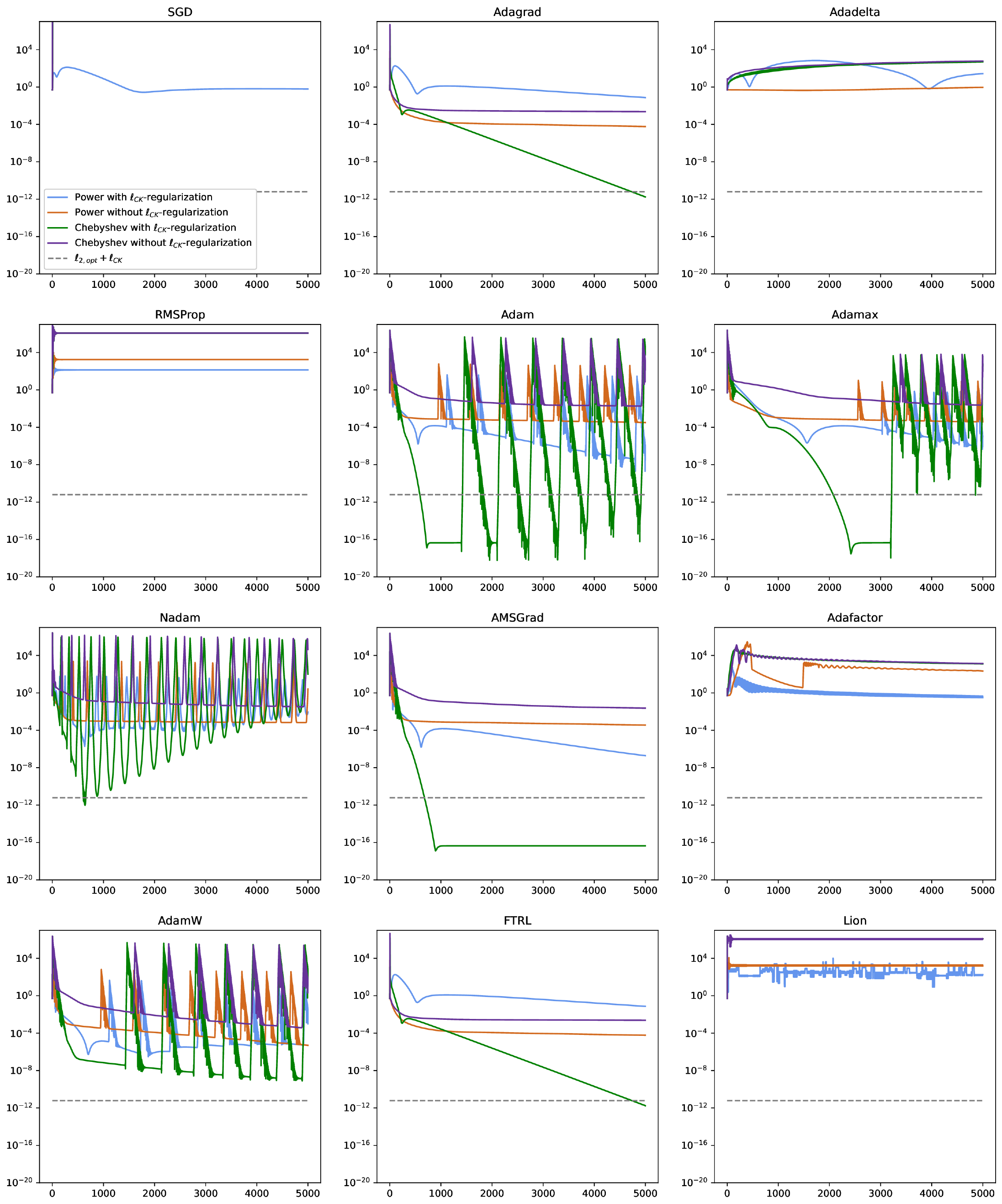}
    \caption{Losses over epochs with different optimizers with learning rate =
    $0.1$, $\alpha = 0.1$, dataset A, $2$ segments. Dashed lines denote $\ell_2
    + \ell_{CK}$ for the respective segment-wise least squares optimum.}
    \label{fig:optimizers}
  \end{figure}

\end{subappendices}

\end{document}